\documentclass[conference]{IEEEtran}
\IEEEoverridecommandlockouts
\usepackage{cite}
\usepackage{amsmath,amssymb,amsfonts}
\usepackage{graphicx}
\usepackage{textcomp}
\usepackage{xcolor}
\usepackage{amsmath}
\usepackage{adjustbox}
\usepackage{algorithm}
\usepackage{algpseudocode}
\usepackage{enumitem}
\usepackage{booktabs}
\usepackage{array}
\usepackage{lipsum}

\newcolumntype{L}[1]{>{\centering\arraybackslash}p{#1}}

\def\BibTeX{{\rm B\kern-.05em{\sc i\kern-.025em b}\kern-.08em
    T\kern-.1667em\lower.7ex\hbox{E}\kern-.125emX}}
\begin{document}

\title{Enhancing Financial Fraud Detection with Human-in-the-Loop Feedback and Feedback Propagation}

\author{\IEEEauthorblockN{Prashank Kadam}
\IEEEauthorblockA{\textit{Vesta Corporation} \\
Atlanta, US \\
prashank.kadam@vesta.io}
}

\maketitle

\begin{abstract}

Human-in-the-loop (HITL) feedback mechanisms can significantly enhance machine learning models, particularly in financial fraud detection, where fraud patterns change rapidly, and fraudulent nodes are sparse. Even small amounts of feedback from Subject Matter Experts (SMEs) can notably boost model performance. This paper examines the impact of HITL feedback on both traditional and advanced techniques using proprietary and publicly available datasets. Our results show that HITL feedback improves model accuracy, with graph-based techniques benefiting the most. We also introduce a novel feedback propagation method that extends feedback across the dataset, further enhancing detection accuracy. By leveraging human expertise, this approach addresses challenges related to evolving fraud patterns, data sparsity, and model interpretability, ultimately improving model robustness and streamlining the annotation process.
\end{abstract}


\section{Introduction}

Financial fraud detection is essential for maintaining the security and integrity of financial systems. As fraud techniques become more sophisticated, traditional detection methods struggle to effectively identify and prevent fraud. Machine learning (ML) techniques have emerged as powerful tools, leveraging large datasets to detect patterns and anomalies indicative of fraud. However, these systems face challenges such as the need for extensive labeled data, the dynamic nature of fraud, and the complexity of domain-specific knowledge. Human-in-the-loop (HITL) feedback mechanisms offer a promising solution to these challenges by incorporating human expertise into the ML process.

HITL involves active human participation in the machine learning pipeline, providing critical insights, annotations, and feedback to enhance model performance. It addresses key issues such as limited labeled data, model interpretability, and adapting to evolving fraud patterns. In financial fraud detection, HITL systems leverage domain knowledge to identify subtle patterns that automated models might overlook. This allows for more accurate model training and validation, reducing false positives and ensuring better fraud detection.

Fraud detection presents several challenges ideal for HITL, including imbalanced datasets, adversarial fraudsters, and complex fraud patterns. Fraudsters continually adapt to evade detection, creating an environment where models must be updated frequently. Graph-based approaches, which model transactions as networks, have shown promise in capturing complex fraud patterns but require expert optimization.

In this paper, we introduce a HITL framework for financial fraud detection, combining human expertise with advanced ML techniques. Our approach incorporates annotation from proprietary and public datasets, interactive model training, and a novel feedback propagation algorithm. We evaluate the impact of HITL feedback using standard metrics, demonstrating improvements in detection accuracy, robustness, and interpretability.

By integrating HITL into fraud detection systems, we improve data annotation, model interpretability, and adaptability to dynamic fraud patterns. Our proposed framework combines advanced ML techniques with a novel feedback propagation method, significantly enhancing fraud detection performance across various algorithms. This research highlights the potential of HITL to improve both traditional and state-of-the-art methods in financial fraud detection while introducing a novel technique for propagating feedback signals throughout the dataset.

\section{Background and Related Work}
\subsection{Human-in-the-Loop (HITL) Feedback}
Human-in-the-Loop (HITL) systems have become integral to modern machine learning, addressing challenges in data processing and model training by enhancing data quality, model interpretability, and performance \cite{wu2022survey}. HITL systems combine human intuition with machine learning, especially in data annotation where labeled data is scarce. For instance, systems proposed by Zhang et al. \cite{zhang2019entity} and Liu et al. \cite{liu2019person} demonstrate significant performance improvements with human feedback. HITL has also been used to refine models in tasks like question answering \cite{wallace2019trivia} and reading comprehension \cite{bartolo2020reading}.

Beyond annotation, HITL has been applied in domains such as computer vision, NLP, and medical applications. Gentile et al. \cite{gentile2019dictionary} utilized HITL for interactive dictionary expansion, while Liu et al. \cite{liu2019person} explored it in person re-identification. HITL systems handle complex tasks across fields, such as scene categorization \cite{yu2015scene}, syntactic parsing \cite{he2016parser}, network anomaly detection \cite{fan2019anomaly}, and outlier detection \cite{chai2020outlier}. Ristoski et al. \cite{ristoski2020relation} proved HITL's utility in relation extraction.

\subsection{Financial Fraud Detection}

Financial fraud detection has evolved from manual assessments to rule-based detection systems \cite{ann1}. These systems, while easy to implement, lack adaptability to changing fraud patterns. Machine learning addressed these limitations with classifiers like Logistic Regression \cite{log-reg1}, Support Vector Machines \cite{svm1}, and Autoencoders with One-class SVMs \cite{svm4}. KNN-clustering \cite{knn1}, Naive Bayes \cite{bayes1}, and ensemble techniques like Random Forest \cite{rf1} and XGBoost \cite{xg1} further advanced fraud detection.

Graph representation learning significantly improved fraud detection. Techniques like DeepWalk \cite{deepw}, node2vec \cite{n2v}, and LINE \cite{line} evolved to advanced methods like Graph Convolutional Networks (GCNs) \cite{gcn} and Graph Attention Networks (GAT) \cite{gat}. Semi-supervised methods \cite{semignn}, sampling-based GNNs \cite{pnc}, and Care-GNN \cite{caregnn} have addressed fraud detection challenges. Recent techniques like Split-GNN \cite{splitgnn}, BOLT \cite{bolt}, and RioGNN \cite{rio} demonstrated state-of-the-art results. GTAN \cite{gtan} further improved results with a Gated Temporal Attention mechanism.

Despite these advancements, challenges like adapting to dynamic fraud patterns and data sparsity remain, which could be mitigated by HITL.

\section{Method}

\subsection{Manual Annotation}

Manual annotation is a key part of our HITL framework, involving experts with domain knowledge in financial fraud annotating both proprietary and publicly available datasets. The annotation process consists of:

\textit{Data Preprocessing:} Data is cleaned by standardizing formats, handling missing values, and performing exploratory analysis to understand feature distributions.

\textit{Annotation Guidelines:} Clear instructions are provided to ensure consistency, defining fraudulent and non-fraudulent transactions. The $isFraud$ label is added to each anchor (e.g., transaction or review) and scaled from 0 to 100 based on SME feedback.

\textit{Quality Control:} A multi-layered process ensures high-quality annotations through cross-checking, discussion, and periodic review to correct errors.

\subsection{Feedback Propagation}

Due to the infeasibility of manually annotating all transactions, we define an algorithm to propagate the feedback signal through the transaction graph. Typically, only 0.1-0.2\% of transactions can be annotated, so propagation extends human feedback to the broader dataset.

\subsubsection{Graph Construction}

We construct a transaction graph $G$, where each node represents a transaction with attributes such as email, phone number, and payment details. Nodes are connected if they share attributes, with edge weights corresponding to the number of shared attributes and their predefined importance (e.g., a phone number carries more weight than an email in telecom datasets).

The edge weight $W_{ij}$ between two transaction nodes $i$ and $j$ can be calculated as follows:

\begin{equation}
W_{ij} = \sum_{k=1}^m w_k \cdot \delta(a_i^k, a_j^k)
\label{eq:weight}
\end{equation}

where $m$ is the number of attributes. $w_k$ is the predefined weight of the $k$-th attribute. $a_i^k$ and $a_j^k$ are the $k$-th attributes of transactions $i$ and $j$, respectively. $\delta(a_i^k, a_j^k)$ is an indicator function that equals 1 if $a_i^k = a_j^k$ and 0 otherwise.

Thus, the transaction graph $G$ can be defined as $G = (V, E, W)$, where $V$ is the set of transaction nodes. $E$ is the set of edges between nodes. $W$ is the set of edge weights.

\begin{figure}[t]
    \centering
    \includegraphics[width=40mm]{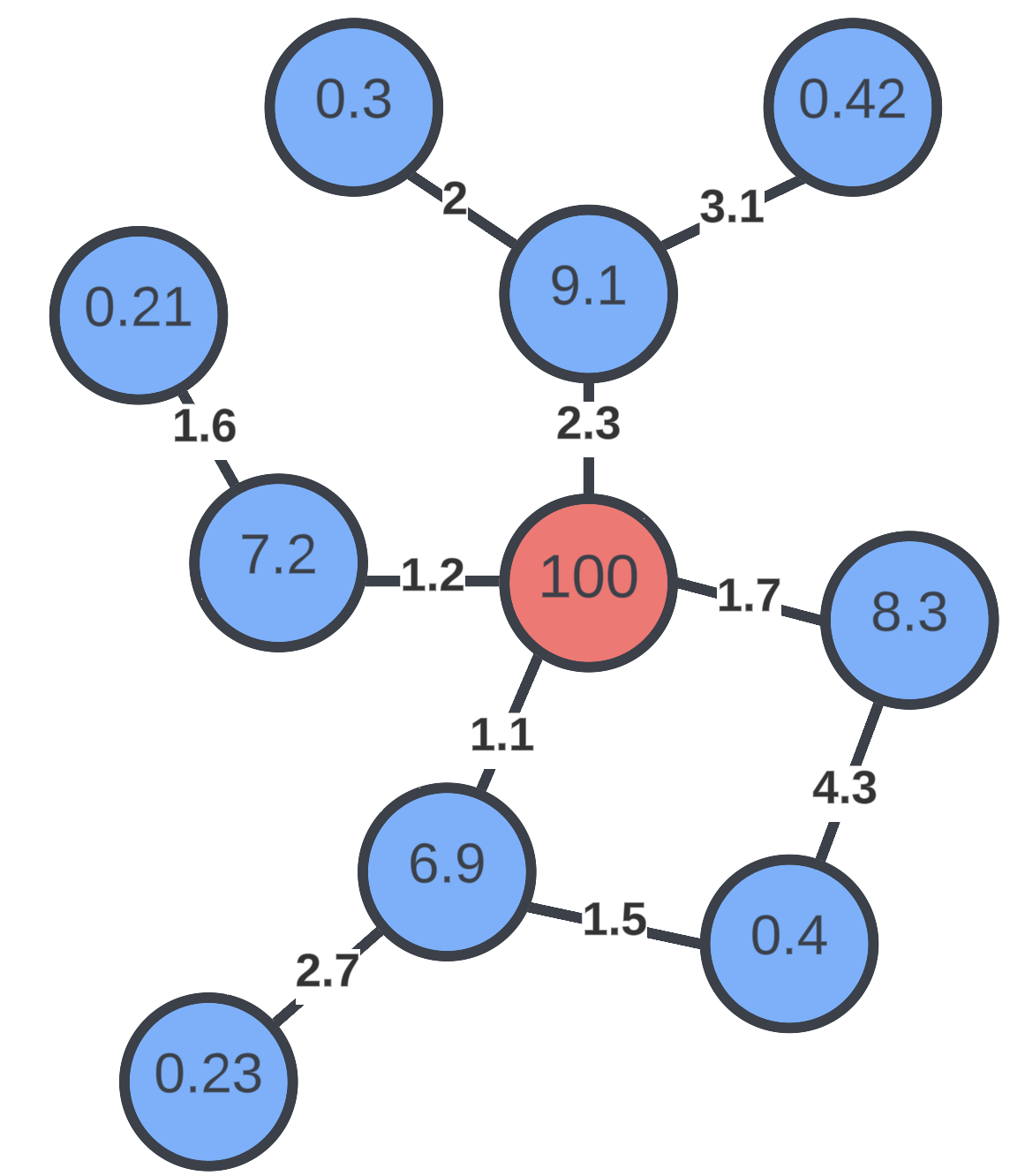}
    \caption{Transaction Graph - Here the red node has been manually annotated where the $isFraud$ score has been set to 100. You can see here how the score is discounted based on the edge weights and the node similarities and propogated further through the graph}
    \label{fig:t-graph}
\end{figure}

\subsubsection{Initialization}

To begin with, the transactions that have been annotated as fraudulent by Subject Matter Experts (SMEs) are given an $isFraud$ score $S_i = 100$ where $S_i \in [0, 100]$. This score is different from the actual target label and is stored as a property on the transaction node.

\subsubsection{Propagation}
In this phase, all the neighboring nodes of the annotated nodes are scored iteratively up to $n$ hops. The score is further discounted as we move away from the originally annotated nodes and the propagation is stopped at $n$ hops when the highest change in score for a given step of propagation falls below the convergence criterion $\epsilon$ (usually set as a lower value). The score of the immediate neighbor $j$ of a node $i$ after one step of feedback propagation can be given by:

\begin{equation}
S_j^{(h)} = S_j^{(h)} + S_i^{(h-1)} \times \frac{W_{ij}}{\max(W)} \times \text{Sim}(i, j)
\label{eq:2}
\end{equation}

where $S_j^{(h)}$ is the score of node $j$ at hop $h$. $S_i^{(h-1)}$ is the score of node $i$ at hop $h-1$. $W_{ij}$ is the edge weight between nodes $i$ and $j$. $\max(W)$ is the maximum possible edge weight in the graph. $\text{Sim}(i, j)$ is the similarity between nodes $i$ and $j$.

Once the highest change in score for a given step of propagation falls below the convergence criterion, the propagation is stopped:

\begin{equation}
\Delta S_{\max}^{(h)} < \epsilon
\label{eq:convergence}
\end{equation}

where $\Delta S_{\max}^{(h)}$ is the highest change in score at hop $h$ between two distinct nodes. $\epsilon$ is the convergence criterion.

\begin{algorithm}[h]
\caption{Feedback Propagation Algorithm}
\begin{algorithmic}[1]
\label{alg:feedback_propagation}
\Require Transaction graph $G = (V, E, W)$, Initial scores $S_i$ for annotated nodes, Maximum number of hops $n$, Convergence criterion $\epsilon$
\Ensure Updated scores $S_i$ for all nodes in $G$
\State Initialize $isFraud$ scores $S_i$ by manual annotation
\State Set $h = 1$
\While{$h \leq n$}
    \State $\Delta S_{\max}^{(h)} = 0$
    \For{each edge $(i, j) \in E$}
        \If{$S_i^{(h-1)} > 0$}
            \State Calculate $S_j^{(h)}$ using Equation \ref{eq:2}
            \State $\Delta S = |S_j^{(h)} - S_j^{(h-1)}|$
            \If{$\Delta S > \Delta S_{\max}^{(h)}$}
                \State $\Delta S_{\max}^{(h)} = \Delta S$
            \EndIf
        \EndIf
    \EndFor
    \If{$\Delta S_{\max}^{(h)} < \epsilon$}
        \State \textbf{break}
    \EndIf
    \State $h = h + 1$
\EndWhile
\State \Return $S_i$ for all nodes
\end{algorithmic}
\end{algorithm}

\section{Experiments}

\subsection{Datasets}

\subsubsection{PFFD} - Our Proprietary Financial Fraud Dataset (PFFD) consists of 1.25 million transactions sourced from e-commerce clients. Each transaction includes hashed details such as Name, Address, Email, Phone, Device, and Payment data, ensuring privacy. The target label, Fraud Score (fs), ranges from 0 to 1000, with 1.07\% of the data labeled as fraudulent. The $isFraud$ label, scaled from 1 to 100, serves as a feedback signal, annotated by SMEs. The graph has an average node degree of 14.71 with around 29,000 hypernodes.

\subsubsection{Yelp Dataset} - We also tested our approach on the Yelp-Fraud dataset \cite{caregnn}. Each node represents a review, and attributes like product and user ID are stored as properties. We modeled this graph similarly to the PFFD, enabling us to apply the same techniques to detect fraudulent reviews.

\subsection{Training and Evaluation}

SMEs annotated 3457 nodes with $isFraud$ labels in the PFFD dataset. Cosine Similarity \cite{cosine} was used as the similarity measure during feedback propagation to evaluate connections between transactions. Training was done in mini-batches, with a 60:20:20 train, validation, and test split. During training, each mini-batch was used to create sub-graphs, which were fed into both tabular and graph ML algorithms. Regularization techniques like dropout and L2 were applied to prevent overfitting. Hyperparameter tuning was conducted during validation, and testing was done chronologically across 10 equal parts. After each test set, SMEs provided additional annotations, and feedback propagation was applied to update node scores.

\subsection{Baselines}

We compared HITL with several baselines, including Logistic Regression \cite{log-reg1}, SVM \cite{svm1}, Random Forest \cite{rf1}, and advanced GNN models such as GCN \cite{gcn}, RGCN \cite{rgcn}, and GAT \cite{gat}. We also compared with state-of-the-art methods like CareGNN \cite{caregnn}, Semi-GNN \cite{semignn}, and BOLT \cite{bolt}, ensuring a comprehensive evaluation across tabular and graph-based algorithms.

\subsection{Experimental Setup}

Data was loaded into a Neo4j graph database, with subgraphs extracted for mini-batches using Cypher queries. The training, validation, and test sets were sequentially loaded to avoid future data leakage. Implementations from PyTorch Geometric were used for GNNs, and sklearn \cite{sklearn} was used for tabular methods. All training was conducted on the Databricks platform with NC12 v3 instances supporting 2 NVIDIA Tesla V100 GPUs.

To ensure fair evaluation, the test set was divided into 10 equal parts, with 150 transactions from each class randomly sampled for annotation by SMEs. After annotation, feedback propagation was performed across $n$ hops, and subsequent evaluations were triggered.

\textit{Evaluation Metrics:} Given the imbalanced nature of the data, we used ROC-AUC and Recall to measure model performance.

\subsection{Ablation Studies}

We performed ablation studies across three different conditions:

\begin{itemize}
    \item Models without Feedback
    \item Models with HITL Feedback
    \item Models with HITL Feedback and Feedback Propagation
\end{itemize}

The results in Table \ref{tab:results} reflect performance across these scenarios, helping quantify the contributions of HITL and feedback propagation in enhancing model accuracy.

\section{Results}

\subsection{Overall Performance Improvement}
All the algorithms perform better with the inclusion of feedback and show further improvement with feedback propagation. Even a small number of annotated transactions, when aided with feedback propagation, were able to catch changing patterns in fraud over time. The inclusion of human annotations results in higher quality training data, leading to improved model performance. Models trained with annotated data exhibit better generalization to unseen fraud patterns and lower false positive rates. Table \ref{tab:results} shows the AUC and Recall values for each of the algorithms without HITL, with HITL, and with HITL and Feedback Propagation. For the PFFD dataset, the average improvement in AUC from without HITL is 7.24\%, and further from HITL to HITL with Feedback Propagation is 2.19\%. For Recall, the improvements are 6.81\% and 2.33\% respectively. For the Yelp fraud dataset, the improvements in AUC are 5.32\% and 2.07\%, and in Recall are 4.31\% and 2.81\%.

\subsection{Tabular vs. Graph Algorithms}
Tabular algorithms show the least amount of improvement with feedback, while graph algorithms benefit considerably. This difference is due to the interactions with the immediate neighbors that graph algorithms take into consideration, allowing them to better capture complex relationships between transactions. For tabular algorithms on the PFFD dataset, the average improvement in AUC from “without feedback” (w/o FB) to “with feedback” (w/ FB) is 3.46\%, while the improvement from w/ FB to “with feedback propagation” (w/ FP) is 2.72\%. Similarly, the average improvement in Recall from w/o FB to w/ FB is 2.83\%, and from w/ FB to w/ FP is 1.64\%. In the case of graph algorithms on the PFFD dataset, the average improvement in AUC from w/o FB to w/ FB is 9.06\%, and from w/ FB to w/ FP is 2.21\%. The average improvement in Recall from w/o FB to w/ FB is 9.09\%, and from w/ FB to w/ FP is 2.59\%. For tabular algorithms on the Yelp dataset, the average improvement in AUC from w/o FB to w/ FB is 3.71\%, and from w/ FB to w/ FP is 2.05\%. The average improvement in Recall from w/o FB to w/ FB is 2.37\%, and from w/ FB to w/ FP is 2.66\%. For graph algorithms on the Yelp dataset, the average improvement in AUC from w/o FB to w/ FB is 6.11\%, and from w/ FB to w/ FP is 1.94\%. The average improvement in Recall from w/o FB to w/ FB is 4.99\%, and from w/ FB to w/ FP is 3.14\%.

\subsection{Progressive Improvement During Testing}
\begin{figure*}[t]
    \includegraphics[width=180mm]{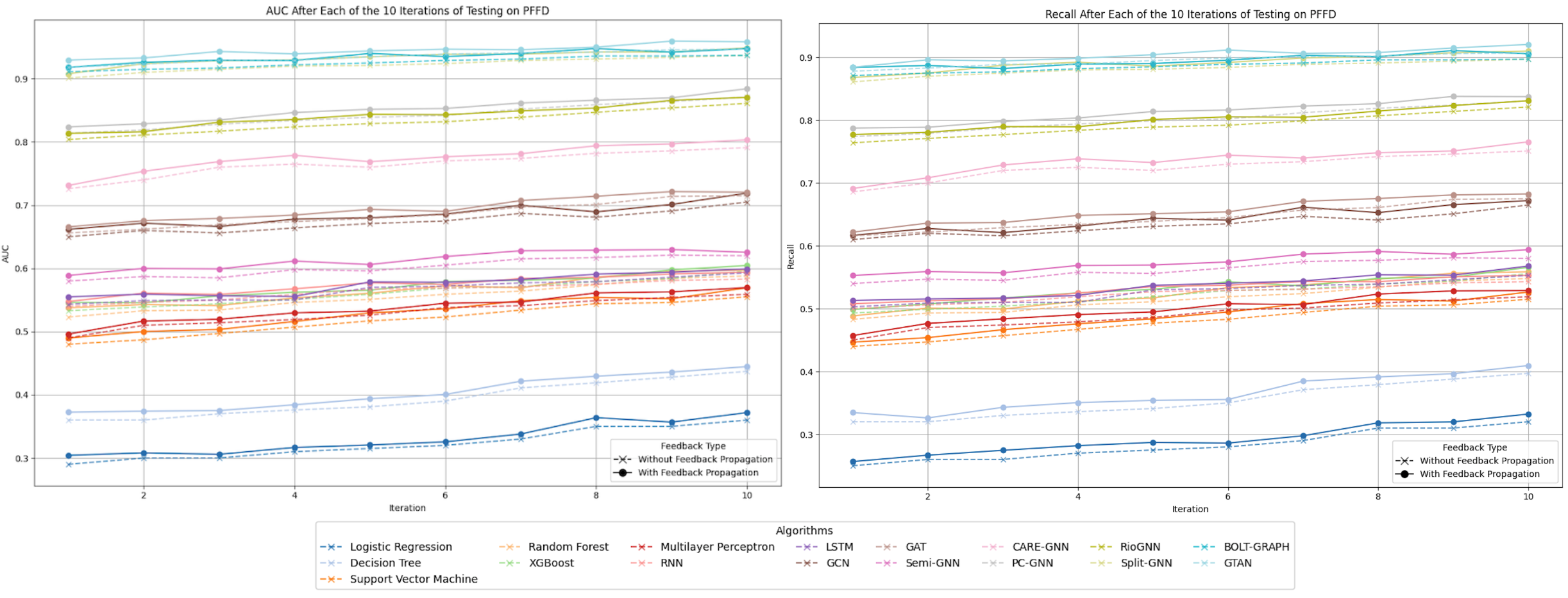}
    \caption{AUC and Recall plots for Test Iterations of the PFFD dataset with and without Feedback Propagation}
    \label{fig:image1}
\end{figure*}

After each test set and feedback propagation cycle during testing, the performance of almost all algorithms improves further. This progressive improvement demonstrates the dynamic adaptability of the models in response to newly annotated data and propagated feedback signals. Figure \ref{fig:image1} shows progressive improvements in the performance of all the evaluated algorithms without HITL, with HITL, and with HITL and Feedback Propagation.

\subsection{Ablation Study Findings}
Ablation studies reveal that both human annotations and feedback propagation are crucial for achieving the best performance. Removing either component results in a significant drop in accuracy and other performance metrics. For instance, in the PFFD dataset, the average improvement in AUC from without HITL to HITL is 7.24\%, and further from HITL to HITL with Feedback Propagation is 2.19\%. The Recall improvements for the same transitions are 6.81\% and 2.33\% respectively. In the Yelp dataset, the AUC improvements from without HITL to HITL and from HITL to HITL with Feedback Propagation are 5.32\% and 2.07\% respectively, while the Recall improvements are 4.31\% and 2.81\% respectively. This underscores the importance of combining human expertise with advanced machine learning techniques to develop robust fraud detection systems.

These results validate the proposed HITL framework and highlight the potential of integrating human expertise into machine learning models for financial fraud detection. The combination of high-quality annotations and effective feedback propagation leads to robust, adaptable, and interpretable fraud detection systems.

% First table with all algorithms
\begin{table*}[t]
\centering
\caption{Averaged Experimental Results for PFFD and Yelp Datasets}
\resizebox{\textwidth}{!}{%
\begin{tabular}{|L{1.4cm}|L{1.1cm}|L{1.1cm}|L{1.1cm}|L{1.1cm}|L{1.1cm}|L{1.1cm}|L{1.1cm}|L{1.1cm}|L{1.1cm}|L{1.1cm}|L{1.1cm}|L{1.1cm}|}
\hline
\textbf{Algorithm} & \multicolumn{6}{c|}{\textbf{PFFD Dataset}} & \multicolumn{6}{c|}{\textbf{Yelp Dataset}} \\ \cline{2-13} 
 & \textbf{AUC (w/o FB)} & \textbf{AUC (w/ FB)} & \textbf{\textbf{AUC (w/ FP)}} & \textbf{Recall (w/o FB)} & \textbf{Recall (w/ FB)} & \textbf{\textbf{Recall (w/ FP)}} & \textbf{AUC (w/o FB)} & \textbf{AUC (w/ FB)} & \textbf{\textbf{AUC (w/ FP)}} & \textbf{Recall (w/o FB)} & \textbf{Recall (w/ FB)} & \textbf{\textbf{Recall (w/ FP)}} \\ \hline
Logistic Regression &               0.33 & 0.34 & \textbf{0.35} & 0.32 & 0.33 & \textbf{0.33} & 0.29 & 0.31 & \textbf{0.32} & 0.27 & 0.29 & \textbf{0.30} \\ \hline
Decision Tree &                     0.38 & 0.39 & \textbf{0.41} & 0.39 & 0.41 & \textbf{0.42} & 0.34 & 0.35 & \textbf{0.36} & 0.33 & 0.33 & \textbf{0.34} \\ \hline
Support Vector Machine &            0.51 & 0.55 & \textbf{0.56} & 0.50 & 0.51 & \textbf{0.52} & 0.44 & 0.45 & \textbf{0.45} & 0.42 & 0.42 & \textbf{0.43} \\ \hline
Random Forest &                     0.53 & 0.54 & \textbf{0.55} & 0.51 & 0.52 & \textbf{0.53} & 0.46 & 0.47 & \textbf{0.48} & 0.43 & 0.43 & \textbf{0.44} \\ \hline
XGBoost &                           0.53 & 0.54 & \textbf{0.55} & 0.52 & 0.53 & \textbf{0.54} & 0.47 & 0.49 & \textbf{0.50} & 0.45 & 0.47 & \textbf{0.47} \\ \hline
Multilayer Perceptron &             0.53 & 0.55 & \textbf{0.56} & 0.52 & 0.53 & \textbf{0.55} & 0.52 & 0.54 & \textbf{0.56} & 0.51 & 0.53 & \textbf{0.55} \\ \hline
RNN &                               0.52 & 0.55 & \textbf{0.57} & 0.50 & 0.53 & \textbf{0.54} & 0.53 & 0.56 & \textbf{0.57} & 0.51 & 0.55 & \textbf{0.56} \\ \hline
LSTM &                              0.54 & 0.59 & \textbf{0.59} & 0.53 & 0.56 & \textbf{0.57} & 0.55 & 0.58 & \textbf{0.61} & 0.54 & 0.56 & \textbf{0.57} \\ \hline
GCN &                               0.62 & 0.70 & \textbf{0.72} & 0.61 & 0.68 & \textbf{0.71} & 0.58 & 0.63 & \textbf{0.66} & 0.56 & 0.60 & \textbf{0.64} \\ \hline
GAT &                               0.63 & 0.72 & \textbf{0.73} & 0.61 & 0.71 & \textbf{0.72} & 0.61 & 0.66 & \textbf{0.67} & 0.59 & 0.63 & \textbf{0.65} \\ \hline
Semi-GNN &                          0.57 & 0.62 & \textbf{0.64} & 0.56 & 0.60 & \textbf{0.63} & 0.64 & 0.68 & \textbf{0.70} & 0.62 & 0.67 & \textbf{0.69} \\ \hline
CARE-GNN &                          0.71 & 0.79 & \textbf{0.81} & 0.69 & 0.78 & \textbf{0.79} & 0.73 & 0.78 & \textbf{0.81} & 0.72 & 0.75 & \textbf{0.78} \\ \hline
PC-GNN &                            0.80 & 0.87 & \textbf{0.89} & 0.78 & 0.85 & \textbf{0.88} & 0.85 & 0.88 & \textbf{0.89} & 0.82 & 0.86 & \textbf{0.89} \\ \hline
RioGNN &                            0.81 & 0.86 & \textbf{0.87} & 0.80 & 0.85 & \textbf{0.86} & 0.87 & 0.89 & \textbf{0.90} & 0.86 & 0.87 & \textbf{0.89} \\ \hline
Split-GNN &                         0.87 & 0.93 & \textbf{0.95} & 0.86 & 0.91 & \textbf{0.93} & 0.87 & 0.93 & \textbf{0.94} & 0.85 & 0.87 & \textbf{0.90} \\ \hline
BOLT-GRAPH &                        0.88 & 0.93 & \textbf{0.95} & 0.86 & 0.92 & \textbf{0.94} & 0.89 & 0.94 & \textbf{0.95} & 0.87 & 0.91 & \textbf{0.92} \\ \hline
GTAN &                              0.88 & 0.94 & \textbf{0.96} & 0.87 & 0.92 & \textbf{0.94} & 0.89 & 0.95 & \textbf{0.95} & 0.87 & 0.92 & \textbf{0.93} \\ \hline
\end{tabular}%
}
\label{tab:results}
\end{table*}

\section{Conclusion}

This study demonstrates the significant benefits of integrating human-in-the-loop (HITL) feedback and feedback propagation in financial fraud detection. By incorporating human expertise through manual annotations and propagating this feedback through transaction graphs, we achieve notable improvements in model performance. Our experiments confirm that graph-based models benefit the most from HITL feedback, showing significant gains in precision, recall, and robustness. The progressive improvements with each feedback cycle validate the adaptability of our approach, effectively capturing evolving fraud patterns even with minimal annotations. Ablation studies further emphasize the importance of both human annotations and feedback propagation, with either component’s removal leading to a drop in accuracy. In conclusion, the proposed HITL framework offers a robust and interpretable solution for financial fraud detection, showcasing the powerful synergy between human expertise and machine learning. Future work could extend this framework to other domains and explore further enhancements in feedback mechanisms and propagation techniques.


\begin{thebibliography}{00}
\bibitem{wu2022survey} X. Wu, L. Xiao, Y. Sun, J. Zhang, T. Ma, and L. He, ``A survey of human-in-the-loop for machine learning,'' \textit{arXiv preprint arXiv:2108.00941}, 2022.
\bibitem{zhang2019entity} Z. Zhang, M. Zhao, and Y. Li, ``Entity extraction with human-in-the-loop,'' \textit{Proceedings of the 2019 IEEE International Conference on Big Data (Big Data)}, pp. 837-844, 2019. doi: 10.1109/BigData47090.2019.9006278.
\bibitem{liu2019person} W. Liu, Y. Wang, and S. Lian, ``Person re-identification with human-in-the-loop,'' \textit{Pattern Recognition Letters}, vol. 125, pp. 1-8, 2019. doi: 10.1016/j.patrec.2019.04.001.
\bibitem{wallace2019trivia} E. Wallace, S. Feng, N. Kandpal, M. Gardner, and S. Singh, ``Universal adversarial triggers for attacking and analyzing NLP,'' \textit{Proceedings of the 2019 Conference on Empirical Methods in Natural Language Processing and the 9th International Joint Conference on Natural Language Processing (EMNLP-IJCNLP)}, pp. 2153-2162, 2019. doi: 10.18653/v1/D19-1221.
\bibitem{bartolo2020reading} M. Bartolo, A. Roberts, S. Riedel, and P. Stenetorp, ``Beat the AI: Investigating adversarial human annotation for reading comprehension,'' \textit{Transactions of the Association for Computational Linguistics}, vol. 8, pp. 662-678, 2020. doi: 10.1162/tacla00342.
\bibitem{gentile2019dictionary} A. Gentile, Z. Zhang, and J. Z. Pan, ``Interactive dictionary expansion using human-in-the-loop,'' \textit{Proceedings of the 2019 Conference on Empirical Methods in Natural Language Processing and the 9th International Joint Conference on Natural Language Processing (EMNLP-IJCNLP)}, pp. 3681-3691, 2019. doi: 10.18653/v1/D19-1374.
\bibitem{yu2015scene} F. Yu, V. Koltun, and T. Funkhouser, ``Scene and object categories with human-in-the-loop,'' \textit{Proceedings of the IEEE Conference on Computer Vision and Pattern Recognition (CVPR)}, pp. 3172-3180, 2015. doi: 10.1109/CVPR.2015.7298962.
\bibitem{he2016parser} L. He, M. Lewis, and L. Zettlemoyer, ``Human-in-the-loop syntactic parsing,'' \textit{Proceedings of the 2016 Conference on Empirical Methods in Natural Language Processing (EMNLP)}, pp. 1237-1247, 2016. doi: 10.18653/v1/D16-1120.
\bibitem{fan2019anomaly} X. Fan, M. Liu, and Z. Lin, ``Human-in-the-loop for network anomaly detection,'' \textit{Proceedings of the 28th ACM International Conference on Information and Knowledge Management (CIKM)}, pp. 1753-1761, 2019. doi: 10.1145/3357384.3357981.
\bibitem{chai2020outlier} S. Chai, Y. Duan, and H. Li, ``Outlier detection with human-in-the-loop,'' \textit{Proceedings of the 29th ACM International Conference on Information and Knowledge Management (CIKM)}, pp. 2417-2420, 2020. doi: 10.1145/3340531.3411960.
\bibitem{ristoski2020relation} P. Ristoski, C. Bizer, and H. Paulheim, ``Human-in-the-loop for relation extraction,'' \textit{Journal of Web Semantics}, vol. 65, p. 100602, 2020. doi: 10.1016/j.websem.2020.100602.
\bibitem{juniper} C. Malone, ``Online Payment Fraud,'' Juniper Research, 2023. [Online]. Available: https://www.juniperresearch.com/research/fintech-payments/fraud-identity/online-payment-fraud-research-report/. [Accessed: Jul. 26, 2024].
\bibitem{rule-based} J. West, M. Bhattacharya, and R. Islam, ``Intelligent Financial Fraud Detection Practices: An Investigation,'' in \textit{10th International ICST Conference on Security and Privacy in Communication Networks (SecureComm)}, Beijing, China, Sep. 2014, pp. 186-203. doi: 10.1007/978-3-319-23802-9-16.
\bibitem{log-reg1} P. Ravisankar, V. Ravi, G. R. Rao, and I. Bose, ``Detection of financial statement fraud and feature selection using data mining techniques,'' \textit{Decision Support Systems}, vol. 50, no. 2, pp. 491-500, 2011. doi: https://doi.org/10.1016/j.dss.2010.11.006.
\bibitem{log-reg2} S. Chen, Y. Goo, and Z. Shen, ``A Hybrid Approach of Stepwise Regression, Logistic Regression, Support Vector Machine, and Decision Tree for Forecasting Fraudulent Financial Statements,'' \textit{TheScientificWorldJournal}, vol. 2014, Sep. 2014. doi: 10.1155/2014/968712.
\bibitem{svm1} N. K. Gyamfi and J. Abdulai, ``Bank Fraud Detection Using Support Vector Machine,'' in \textit{2018 IEEE 9th Annual Information Technology, Electronics and Mobile Communication Conference (IEMCON)}, 2018, pp. 37-41. doi: 10.1109/IEMCON.2018.8614994.
\bibitem{svm2} G. G. Sundarkumar, V. Ravi, and V. Siddeshwar, ``One-class support vector machine based undersampling: Application to churn prediction and insurance fraud detection,'' in \textit{2015 IEEE International Conference on Computational Intelligence and Computing Research (ICCIC)}, 2015, pp. 1-7. doi: 10.1109/ICCIC.2015.7435726.
\bibitem{svm3} X. Li and S. Ying, ``Lib-SVMs Detection Model of Regulating-Profits Financial Statement Fraud Using Data of Chinese Listed Companies,'' in \textit{2010 International Conference on Electrical and Electronics Engineering (ICEEE)}, 2010. doi: 10.1109/ICEEE.2010.5660371.
\bibitem{svm4} M. Jeragh and M. AlSulaimi, ``Combining Auto Encoders and One Class Support Vectors Machine for Fraudulent Credit Card Transactions Detection,'' in \textit{2018 Second World Conference on Smart Trends in Systems, Security and Sustainability (WorldS4)}, 2018, pp. 178-184. doi: 10.1109/WorldS4.2018.8611624.
\bibitem{knn1} J. O. Awoyemi, A. O. Adetunmbi, and S. A. Oluwadare, ``Credit card fraud detection using machine learning techniques: A comparative analysis,'' in \textit{2017 International Conference on Computing Networking and Informatics (ICCNI)}, 2017, pp. 1-9. doi: 10.1109/ICCNI.2017.8123782.
\bibitem{knn2} N. Malini and M. Pushpa, ``Analysis on credit card fraud identification techniques based on KNN and outlier detection,'' in \textit{2017 Third International Conference on Advances in Electrical, Electronics, Information, Communication and Bio-Informatics (AEEICB)}, 2017, pp. 255-258. doi: 10.1109/AEEICB.2017.7972424.
\bibitem{bayes1} Q. Deng, ``Detection of fraudulent financial statements based on Naïve Bayes classifier,'' in \textit{2010 5th International Conference on Computer Science \& Education}, 2010, pp. 1032-1035. doi: 10.1109/ICCSE.2010.5593407.
\bibitem{bayes2} P. Hajek and R. Henriques, ``Mining corporate annual reports for intelligent detection of financial statement fraud – A comparative study of machine learning methods,'' \textit{Knowledge-Based Systems}, vol. 128, pp. 139-152, 2017. doi: https://doi.org/10.1016/j.knosys.2017.05.001.
\bibitem{rf1} S. Xuan, G. Liu, Z. Li, L. Zheng, S. Wang, and C. Jiang, ``Random forest for credit card fraud detection,'' in \textit{2018 IEEE 15th International Conference on Networking, Sensing and Control (ICNSC)}, 2018, pp. 1-6. doi: 10.1109/ICNSC.2018.8361343.
\bibitem{xg1} P. Hájek, M. Abedin, and U. Sivarajah, ``Fraud Detection in Mobile Payment Systems using an XGBoost-based Framework,'' \textit{Information Systems Frontiers}, vol. 25, pp. 1-19, Oct. 2022. doi: 10.1007/s10796-022-10346-6.
\bibitem{ann1} A. Ali, S. A. Razak, S. H. Othman, T. A. E. Eisa, A. Al-Dhaqm, M. Nasser, T. Elshafie, and A. Saif, ``Financial Fraud Detection Based on Machine Learning: A Systematic Literature Review,'' \textit{Applied Sciences}, vol. 12, no. 19, article 9637, 2022. doi: 10.3390/app12199637.
\bibitem{deepw} B. Perozzi, R. Al-Rfou, and S. Skiena, ``DeepWalk: Online Learning of Social Representations,'' in \textit{20th ACM SIGKDD International Conference on Knowledge Discovery and Data Mining}, 2014, pp. 701-710.
\bibitem{n2v} A. Grover and J. Leskovec, ``node2vec: Scalable Feature Learning for Networks,'' in \textit{22nd ACM SIGKDD International Conference on Knowledge Discovery and Data Mining}, 2016, pp. 855-864.
\bibitem{line} J. Tang, M. Qu, M. Wang, J. Zhang, J. Yan, and Q. Mei, ``LINE: Large-scale Information Network Embedding,'' in \textit{24th International Conference on World Wide Web}, 2015, pp. 1067-1077.
\bibitem{sdne} D. Wang, P. Cui, and W. Zhu, ``Structural Deep Network Embedding,'' in \textit{22nd ACM SIGKDD International Conference on Knowledge Discovery and Data Mining}, 2016, pp. 1225-1234.
\bibitem{gcn} T. N. Kipf and M. Welling, ``Semi-Supervised Classification with Graph Convolutional Networks,'' in \textit{5th International Conference on Learning Representations (ICLR)}, Toulon, France, Apr. 2017. [Online]. Available: https://openreview.net/forum?id=SJU4ayYgl.
\bibitem{semignn} D. Wang, Y. Qi, J. Lin, P. Cui, Q. Jia, Z. Wang, Y. Fang, Q. Yu, J. Zhou, and S. Yang, ``A Semi-Supervised Graph Attentive Network for Financial Fraud Detection,'' in \textit{2019 IEEE International Conference on Data Mining (ICDM)}, 2019, pp. 598-607. doi: 10.1109/ICDM.2019.00070.
\bibitem{gat} P. Veličković, G. Cucurull, A. Casanova, A. Romero, P. Liò, and Y. Bengio, ``Graph Attention Networks,'' in \textit{International Conference on Learning Representations (ICLR)}, 2018.
\bibitem{pnc} Y. Liu, X. Ao, Z. Qin, J. Chi, J. Feng, H. Yang, and Q. He, ``Pick and Choose: A GNN-based Imbalanced Learning Approach for Fraud Detection,'' in \textit{WWW '21: The Web Conference 2021}, Virtual Event / Ljubljana, Slovenia, Apr. 2021, pp. 3168-3177. doi: 10.1145/3442381.3449989.
\bibitem{caregnn} Y. Dou, Z. Liu, L. Sun, Y. Deng, H. Peng, and P. S. Yu, ``Enhancing Graph Neural Network-based Fraud Detectors against Camouflaged Fraudsters,'' \textit{Proceedings of the 29th ACM International Conference on Information \& Knowledge Management}, 2020. [Online]. Available: https://api.semanticscholar.org/CorpusID:221186720.
\bibitem{scngnn} J. Chen, Q. Chen, F. Jiang, X. Guo, K. Sha, and Y. Wang, ``SCN GNN: A GNN-based fraud detection algorithm combining strong node and graph topology information,'' \textit{Expert Systems with Applications}, vol. 237, p. 121643, 2024. doi: https://doi.org/10.1016/j.eswa.2023.121643.
\bibitem{sage} W. Hamilton, R. Ying, and J. Leskovec, ``Inductive representation learning on large graphs,'' in \textit{Advances in neural information processing systems}, 2017, pp. 1024-1034.
\bibitem{splitgnn} B. Wu, X. Yao, B. Zhang, K.-M. Chao, and Y. Li, ``SplitGNN: Spectral Graph Neural Network for Fraud Detection against Heterophily,'' in \textit{Proceedings of the 32nd ACM International Conference on Information and Knowledge Management}, New York, NY, USA, 2023, pp. 2737-2746. doi: 10.1145/3583780.3615067.
\bibitem{rgcn} M. Schlichtkrull, T. N. Kipf, P. Bloem, R. van den Berg, I. Titov, and M. Welling, ``Modeling Relational Data with Graph Convolutional Networks,'' in \textit{European Semantic Web Conference}, 2018, pp. 593-607.
\bibitem{mahalanobis} P. C. Mahalanobis, ``On the generalized distance in statistics,'' \textit{Proceedings of the National Institute of Sciences of India}, vol. 2, no. 1, pp. 49-55, 1936.
\bibitem{progdl} H. M. Fayek, L. Cavedon, and H. R. Wu, ``Progressive learning: A deep learning framework for continual learning,'' \textit{Neural Networks}, vol. 128, pp. 345-357, 2020. doi: https://doi.org/10.1016/j.neunet.2020.05.011.
\bibitem{gnn} F. Scarselli, M. Gori, A. C. Tsoi, M. Maggini, and G. Monfardini, ``The Graph Neural Network Model,'' \textit{IEEE Transactions on Neural Networks}, vol. 20, pp. 61-80, 2009.
\bibitem{gsage} W. L. Hamilton, R. Ying, and J. Leskovec, ``Inductive Representation Learning on Large Graphs,'' in \textit{NIPS}, 2017.
\bibitem{barrat} A. Barrat, M. Barthelemy, R. Pastor-Satorras, and A. Vespignani, ``The architecture of complex weighted networks,'' \textit{Proceedings of the National Academy of Sciences}, vol. 101, no. 11, pp. 3747-3752, 2004. doi: 10.1073/pnas.0400087101.
\bibitem{gtan} S. Xiang, M. Zhu, D. Cheng, E. Li, R. Zhao, Y. Ouyang, L. Chen, and Y. Zheng, ``Semi-supervised credit card fraud detection via attribute-driven graph representation,'' in \textit{Proceedings of the Thirty-Seventh AAAI Conference on Artificial Intelligence and Thirty-Fifth Conference on Innovative Applications of Artificial Intelligence and Thirteenth Symposium on Educational Advances in Artificial Intelligence}, 2023. doi: 10.1609/aaai.v37i12.26702.
\bibitem{jaccard} P. Jaccard, “Étude comparative de la distribution florale dans une portion des Alpes et des Jura,” Bulletin de la Société Vaudoise des Sciences Naturelles, vol. 37, pp. 547–579, 1901.
\bibitem{cosine} M. W. Berry and M. Browne, “Understanding Search Engines: Mathematical Modeling and Text Retrieval,” Society for Industrial and Applied Mathematics, 2005.
\bibitem{bolt} N. Meisburger, V. Lakshman, G. Benito, J. Engels, D. T. Ramos, P. Pratik, B. Coleman, B. Meisburger, S. Gupta, Y. Adunukota, S. Jain, T. Medini, and A. Shrivastava, “BOLT: An Automated Deep Learning Framework for Training and Deploying Large-Scale Search and Recommendation Models on Commodity CPU Hardware,” arXiv preprint arXiv:2303.17727, 2023.
\bibitem{rio} H. Peng, R. Zhang, Y. Dou, R. Yang, J. Zhang, and P. S. Yu, "Reinforced Neighborhood Selection Guided Multi-Relational Graph Neural Networks," ACM Transactions on Information Systems, vol. 40, no. 4, article 69, 2022.
\bibitem{rnn} D. E. Rumelhart, G. E. Hinton, and R. J. Williams, “Learning internal representations by error propagation,” Tech. Rep. ICS 8504, Institute for Cognitive Science, University of California, San Diego, 1985
\bibitem{sklearn} F. Pedregosa, G. Varoquaux, A. Gramfort, V. Michel, B. Thirion, O. Grisel, M. Blondel, P. Prettenhofer, R. Weiss, V. Dubourg, J. Vanderplas, A. Passos, D. Cournapeau, M. Brucher, M. Perrot, and E. Duchesnay, "Scikit-learn: Machine Learning in Python," Journal of Machine Learning Research, vol. 12, pp. 2825-2830, 2011.

\end{thebibliography}
\end{document}